\documentclass[11pt,a4paper]{article}
\usepackage[utf8]{inputenc}
\usepackage[T1]{fontenc}
\usepackage{amsmath, amssymb, amsthm, graphicx, float, textcomp}
\usepackage[numbers,sort&compress]{natbib}
\usepackage[colorlinks=true, linkcolor=blue, citecolor=blue, urlcolor=blue]{hyperref}
\usepackage[margin=1in]{geometry}
\usepackage{parskip}
\usepackage{authblk} 

\title{\LARGE \textbf{Cognitive Alignment At No Cost: Inducing Human Attention Biases For Interpretable Vision Transformers}}
\author{Ethan Knights}
\affil{Cambridge, UK}
\date{}

\begin{document}
\maketitle

\section{Abstract}\label{abstract}

For state-of-the-art image understanding, Vision Transformers (ViTs)
have become the standard architecture but their processing diverges
substantially from human attentional characteristics. We investigate
whether this cognitive gap can be shrunk by fine-tuning the
self-attention weights of Google's ViT-B/16 on human saliency fixation
maps. To isolate the effects of semantically relevant signals from
generic human supervision, the tuned model is compared against a
shuffled control. Fine-tuning significantly improved alignment across
five saliency metrics and induced three hallmark human-like biases:
tuning reversed the baseline\textquotesingle s anti-human large-object
bias toward small-objects, amplified the animacy preference and
diminished extreme attention entropy. Bayesian parity analysis provides
decisive to very-strong evidence that this cognitive alignment comes at
no cost to the model's original classification performance on in-
(ImageNet), corrupted (ImageNet-C) and out-of-distribution (ObjectNet)
benchmarks. An equivalent procedure applied to a ResNet-50 Convolutional
Neural Network (CNN) instead degraded both alignment and accuracy,
suggesting that the ViT's modular self-attention mechanism is uniquely
suited for dissociating spatial priority from representational logic.
These findings demonstrate that biologically grounded priors can be
instilled as a free emergent property of human-aligned attention, to
improve transformer interpretability.

\newpage
\section{Introduction}\label{introduction}

As deep neural networks (DNNs) capable of visual understanding are
increasingly deployed in high-stakes settings, questions about how they
arrive at their decisions are becoming as important as how well they
perform. Progress in computer vision has been rapid: Convolutional
Neural Network (CNN) architectures (e.g., AlexNet; Krizhevsky et al.,
2012; ResNet; He et al., 2016) have achieved remarkable accuracy on
large-scale object recognition tasks before being superseded by the
vision transformer (ViT; Dosovitskiy et al., 2021) which now dominates
tasks requiring fusion of vision and language. Across these models are
internal processing strategies that remain somewhat opaque and are
demonstrably non-human.

Architecturally, the design of CNNs and ViTs leads each family to a
unique divergence from human visual attention. For humans, visual
attention is flexible, selective and globally allocated across a scene
from the outset, with information being processed either in
high-resolution at the fovea or coarsely in the periphery (Itti \& Koch,
2001). Despite being loosely inspired by the hierarchical organisation
of the visual cortex (Hubel \& Wiesel, 1962; LeCun et al., 1998), CNNs
instead process images through fixed local receptive fields that expand
only gradually across layers, lacking the flexible scene-wide
prioritisation characteristic of human vision. The patch-embedding and
self-attention mechanisms common to ViTs (Vaswani et al., 2017) lead to
a uniform image patch processing strategy that lacks foveal priority or
salience-driven selectivity. In humans, such parallel processing is
largely confined to the pre-attentive stage (i.e., the rapid automatic
detection of low-level features like colour and orientation; Treisman \&
Gelade, 1980) before feeding into a capacity-limited, serial attentional
system (Duncan, 1984; Itti \& Koch, 2001). Novel architectural
mechanisms are rapidly evolving are continuing to yield remarkable
engineering gains (e.g., vision-language top-down attention; Li et al.,
2020; Radford et al., 2021; Kirillov et al., 2023) but (even in cases
where these do resemble functional human analogues) their biological
equivalence is, at best, oversimplified (Yamins \& DiCarlo, 2016).

Human fixations offer a validated proxy of visual attention (Petersen \&
Posner, 2012) that carry rich theoretically grounded signals that make
them a principled target for both inducing and measuring human-like
processing in models. Despite inherent stochasticity in eye movement
distributions (Krajbich et al., 2010), fixations are not random and are
directed to maximise information acquisition (Hayhoe \& Rothkopf, 2011).
Under free-viewing conditions, salient objects attract attention early
and reliably (Parkhurst et al., 2002), with bottom-up signals being
particularly influential for initial fixations. More broadly, where
humans look is shaped by competition between signals including object
salience, visual acuity constraints (e.g., eccentricity; Peschel \&
Orquin, 2013), colour and contextual clutter (Janiszewski, 1998). Beyond
these low-level factors, fixations also reflect a rich array of
cognitive biases - systematic attentional preferences that are robust,
replicable and grounded in well-established psychological science. These
include (amongst many others) an automatic preferential bias toward
animate entities (New et al., 2007), faces (Valenza et al., 1996;
Kanwisher \& Yovel, 2006), small informationally dense objects (Knights
et al., 2021; 2022), central image bias (Tatler, 2007) and socially
relevant cues such as gaze direction (Frischen et al., 2007).

A popular empirical approach is to extract attention maps from models
using explainable AI (XAI) techniques for CNNs (e.g., GradCAM; Selvaraju
et al., 2017) or ViTs (e.g., attention rollout; Abnar \& Zuidema, 2020)
and comparing these with human fixations. This has begun to characterise
the architectural divergences above, where model attention is unique
spatially (e.g., which regions a model attends to; Piñero et al., 2025)
and/or temporally (e.g., how attention evolves across model layers;
Carrasco et al., 2025). For CNNs, perhaps the clearest demonstration is
the texture bias: CNNs are strongly disposed toward recognising textures
rather than shapes, in direct contrast to humans who rely predominantly
on shape-based representations (Geirhos et al., 2019) which is a result
of training datasets preserving both texture and shape cues
simultaneously, that allows the network to exploit lower-level
statistics in place of principled object representations (Hermann et
al., 2020). For ViTs, alignment with human fixations appears to vary
considerably across heads, layers and training regime (Yamamoto et al.,
2025; Piñero et al., 2025) and studies of their broader correspondence
with biological vision has shown how their alignment, at least in terms
of a preference for animacy, is poorer than found in the 2012 AlexNet
architecture (Diaz \& Alvarez, 2025). This suggests that deliberate,
targeted interventions, rather than only architectural evolution, may be
uniquely positioned to help close the gap between attention
representations between models and humans.

There is steadily growing evidence showing that fine-tuning DNNs on
human derived signals can induce meaningful cognitive alignment. For
CNNs, Geirhos et al. (2019) demonstrated that training on human-like
representations yielded not only greater alignment for shape-based
processing strategies but emergent performance gains, including improved
robustness to image distortions (e.g., via corruptions). For ViTs,
Koorathota et al. (2023) showed that a fixation-based loss improved
steering prediction (left vs. right decisions) accuracy under
uncertainty. Crucially, these results suggest that complex model
architectures can exhibit more human-like processing strategic alignment
and that this can even actively improve downstream capabilities. Other
recent specific human cognitive biases have also been successfully
induced in DNNs directly including aversive responses to spiders (Pegler
et al., 2025) and preferential processing of faces and objects (Lu \&
Jiang, 2026), which altogether suggest that fine-tuning is a broadly
viable mechanism for instilling psychologically motivated priors into
model behaviour.

Here, we test if an open-source state‑of‑the‑art model (Google's
ViT-B/16; Dosovitskiy et al., 2021) can be fine-tuned (using the SALICON
fixation dataset collected over Microsoft's COCO images; Lin et al.,
2014; Jiang et al., 2015; Figure 1) to attend in a more human‑like
fashion. Beyond measuring general saliency metrics (Bylinskii et al.,
2019; Figure 2) we uniquely test if the model acquires hallmark human
cognitive biases (Figure 3) because, if so, this shows a genuine
improvement to model interpretability as its attention maps can be
understood via the same psychological constructs that explain human
gaze. We focus on three cognitive biases that are well-grounded in
psychological science and directly measurable from the dataset: animacy,
object size and attention sparsity. Animacy reflects a deeply rooted
evolutionary prioritisation of living things (New et al., 2007); object
size captures the tension between low-level salience and semantic
information density (Walther \& Koch, 2006; Nuthmann \& Henderson, 2010;
Henderson \& Hayes, 2017); and attention sparsity reflects the
well-established tendency for human fixations to be spatially
concentrated rather than diffuse (Itti \& Baldi, 2009; Bruce \& Tsotsos,
2009). For rigour, we also include an appropriate shuffled control model
(to ensure that any cognitive effects of tuning are from semantic,
rather than generic human supervision, signals), benchmark the model on
downstream performance (to ensure this comes at no cost to its original
capabilities) and repeat the procedure with a ResNet CNN (to assess if
alignment gains are generalisable across DNN architectures).

\begin{figure}[H]\centering\includegraphics[width=0.9\textwidth,keepaspectratio]{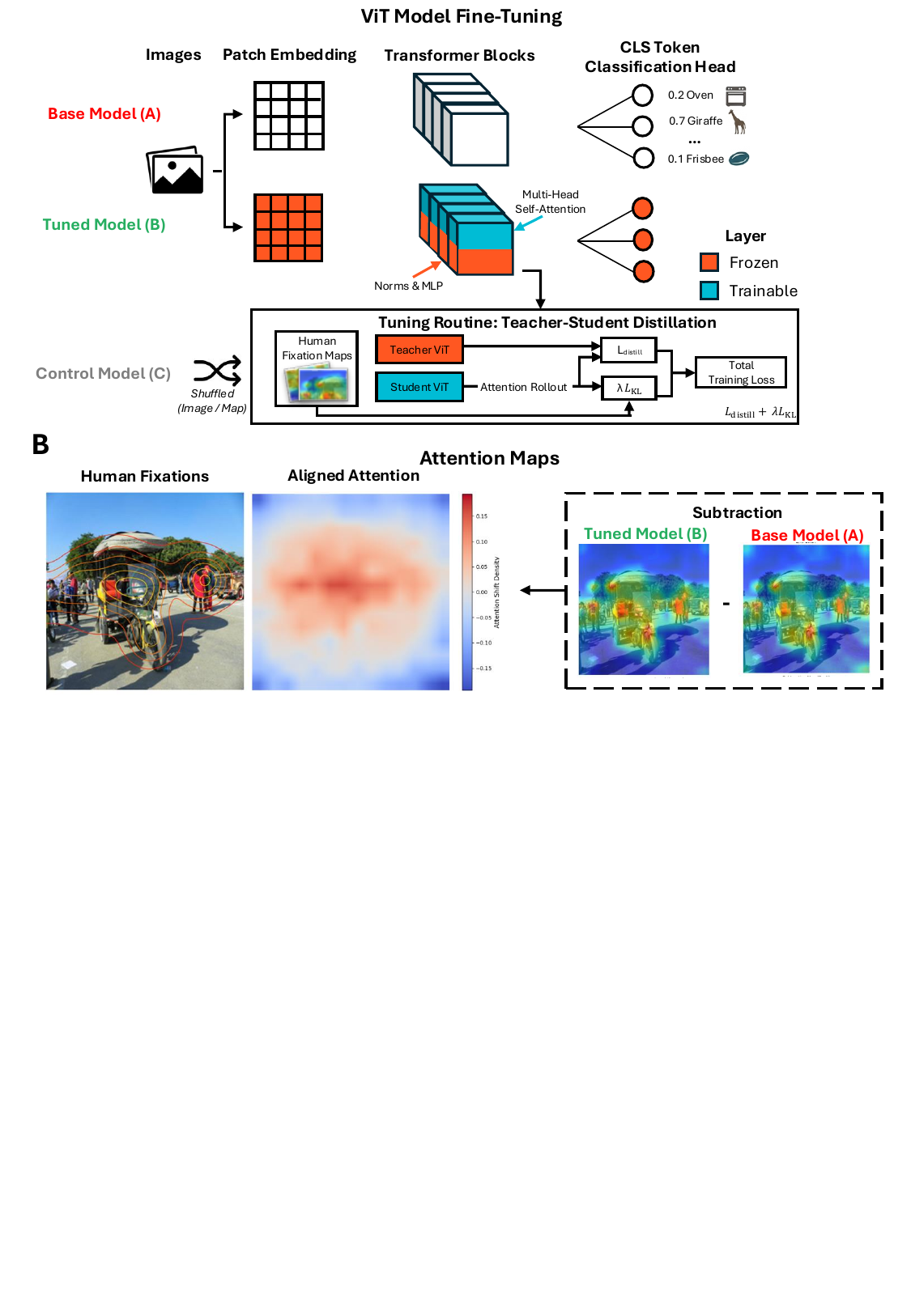}\vspace{-10cm}\end{figure}

\textbf{Figure 1. Methods for ViT Fine-Tuning. (A) Model Variants and
Training Framework.} Three model variants including a baseline frozen
pretrained ViT-B/16 (red - Model A), a tuned human-attention model
(green - Model B) and a control model (grey - Model C). Models B and C
were both trained on human fixation density maps via teacher-student
distillation where only the self-attention weights are updated, but
these maps were mismatched for the control by shuffling. \textbf{(B)
Attention Redistribution.} Human fixations are shown alongside Model
B\textquotesingle s aligned attention map via an attention rollout
difference map (i.e., subtracting Model A\textquotesingle s rollout from
Model B\textquotesingle s) highlighting areas of increased (warm) and
suppressed (cool) attention. Note the redistribution of attention from
uninformative peripheral regions toward central semantically relevant
features.

\section{Results}\label{results}

\subsection{Saliency Alignment}\label{saliency-alignment}

Fine-tuning on human fixation data produced a consistent improvement in
saliency alignment. The tuned model aligned significantly better with
human fixations than both the unmodified baseline and the shuffled
control models across all five metrics (Pearson CC, NSS, AUC-Judd,
KL-divergence, SIM; all comparisons p \textless{} 0.001; Figure 2A).
Hereafter, Pearson CC is used as the primary alignment metric.
Crucially, since the tuned model outperformed the shuffled control - a
model exposed to an equal quantity of human attention supervision, but
with the image-fixation correspondence destroyed, this confirms that the
alignment gain reflects the structured, semantic content of the fixation
signal rather than a generic effect of additional human supervision.

For every image, across every category, alignment gain was positive,
relative to baseline (Figure 2B). The degree of alignment gain was a
function of human fixation consistency where the tuned
model\textquotesingle s improvement over baseline was negatively
correlated with inter-observer consistency (r = -0.55, p \textless{}
0.0001; Figure 2C): gains were largest on images where human annotators
disagreed most about where to look, and smallest where fixation targets
were well-defined and consistent. The observation of diminishing
absolute gain at higher consistency levels is consistent with a headroom
ceiling: as baseline alignment improves on well-defined scenes, the
available margin for further improvement narrows. Although this pattern
could alternatively reflect a tendency to benefit most from ambiguous
images, the positive headroom-normalised gains across all quartiles,
argue against this interpretation.

\begin{figure}[H]\centering\includegraphics[width=0.9\textwidth,keepaspectratio]{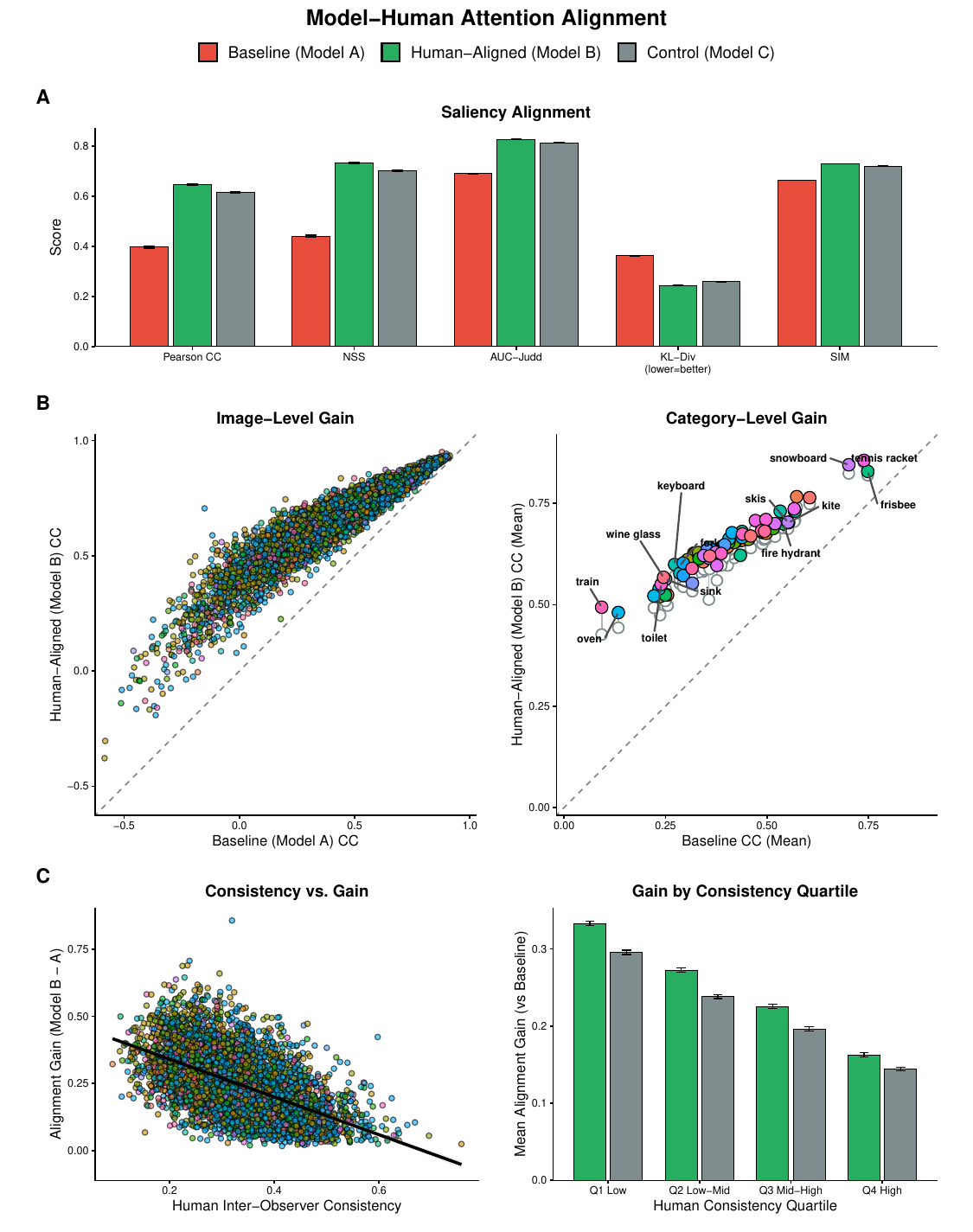}\vspace{-0.5cm}\end{figure}

\textbf{Figure 2. Model-Human Attentional Alignment. (A) Saliency
Alignment.} Mean alignment scores across five saliency metrics for the
baseline (red), tuned (green) and shuffled control (grey) models. Error
bars denote ±1 SEM. \textbf{(B) Alignment Gain.} Left: image-level
scatter of tuned versus baseline Pearson CC, with each point
representing one image; points above the diagonal indicate gains for the
tuned model. Right: category-level alignment, where the filled circles
denote the tuned model mean and their connected grey circles denote the
shuffled control, plotted against the baseline mean. The identity
diagonal represents no gain; no image or category fell below.
\textbf{(C) Gain as a Function of Human Consistency.} Left: scatterplot
of alignment gain (tuned minus baseline Pearson CC) as a function of
human inter-observer consistency, with a least-squares regression line
overlaid (Pearson r = -0.55, p \textless{} 0.0001). Right: mean
alignment gain binned by inter-observer consistency quartile (Q1 =
lowest, Q4 = highest), showing the tuned model\textquotesingle s gain
over the baseline (green) and over the shuffled control (grey).

\subsection{Cognitive Bias}\label{cognitive-bias}

Humans showed the strongest animate preference, based on 2,619 images
containing both animate and inanimate objects, where they allocated
significantly greater attention density to animate regions (animate:
0.45, inanimate: 0.37; $\Delta$ = +0.077, t(2618) = 14.77, p \textless{} 0.001,
d = 0.29). All three models showed a weaker but present bias in the same
direction. The animacy effect was largest for the tuned model (animate:
0.36, inanimate: 0.33; $\Delta$ = +0.029, t(2618) = 9.41, p \textless{} 0.001,
d = 0.18), followed by the shuffled control (animate: 0.34, inanimate:
0.31; $\Delta$ = +0.025, t(2618) = 8.94, p \textless{} 0.001, d = 0.17) and
smallest for the baseline (animate: 0.33, inanimate: 0.30; $\Delta$ = +0.021,
t(2618) = 7.94, p \textless{} 0.001, d = 0.16). Between-model
comparisons confirmed that the tuned model\textquotesingle s animate
attention density was significantly higher than for both the baseline
(t(2618) = 40.14, p \textless{} 0.001, d = 0.78) and the shuffled
control (t(2618) = 42.34, p \textless{} 0.001, d = 0.83), establishing
that the structured fixation signal had specifically amplified the
animacy preference beyond what is produced by generic human supervision.

The object size analysis revealed the sharpest qualitative distinction
across models, conducted across 3,068 images containing both small and
large annotated objects. As expected, humans preferentially attended to
small objects over large ones (small: 0.37, large: 0.34; $\Delta$ = +0.030,
t(3067) = 7.30, p \textless{} 0.001, d = 0.13). The baseline showed the
opposite pattern: a reliable anti-human tendency to over-attend to large
objects (small: 0.26, large: 0.27; $\Delta$ = -0.007, t(3067) = -3.45, p
\textless{} 0.001, d = -0.06). The shuffled control showed no
directional preference (small: 0.28, large = 0.28; $\Delta$ = -0.0004, t(3067)
= -0.21, p = 0.836, d \textless{} 0.01), indicating that generic
supervision neither fully corrects nor amplifies the baseline tendency.
Crucially, the tuned model reversed the baseline\textquotesingle s
direction entirely, showing a reliable small-object preference (small:
0.30, large: 0.29; $\Delta$ = +0.009, t(3067) = 3.96, p \textless{} 0.001, d =
0.07), and between-model comparisons showed its small-object density was
significantly higher than the baseline (t(3067) = 62.06, p \textless{}
0.001, d = 1.12) and shuffled control (t(3067) = 78.22, p \textless{}
0.001, d = 1.41).

Lastly, fine-tuning also produced a modest but reliable reduction in
attention entropy across all 5,000 images. The tuned model was
significantly less entropic than the baseline (15.38 ± 0.12 vs. 15.42 ±
0.09 bits; $\Delta$ = -0.043 bits, t(4999) = -61.90, p \textless{} 0.001, d =
-0.88), indicating a shift toward more spatially concentrated attention.
The shuffled control also showed entropy reduction relative to baseline
($\Delta$ = -0.012 bits, d = -0.31), though smaller in magnitude, again
suggesting the structured signal amplifies an effect that generic
supervision partially produces. All models remained substantially more
entropic than human fixation maps (human: 14.99 ± 0.17 bits; all
model-to-human comparisons; p \textless{} 0.001, d \textgreater{} 2.6),
confirming that fine-tuning moves the distribution in the human-like
direction, without closing the gap to biological levels of attentional
concentration.

\begin{figure}[H]\centering\includegraphics[width=0.9\textwidth,keepaspectratio]{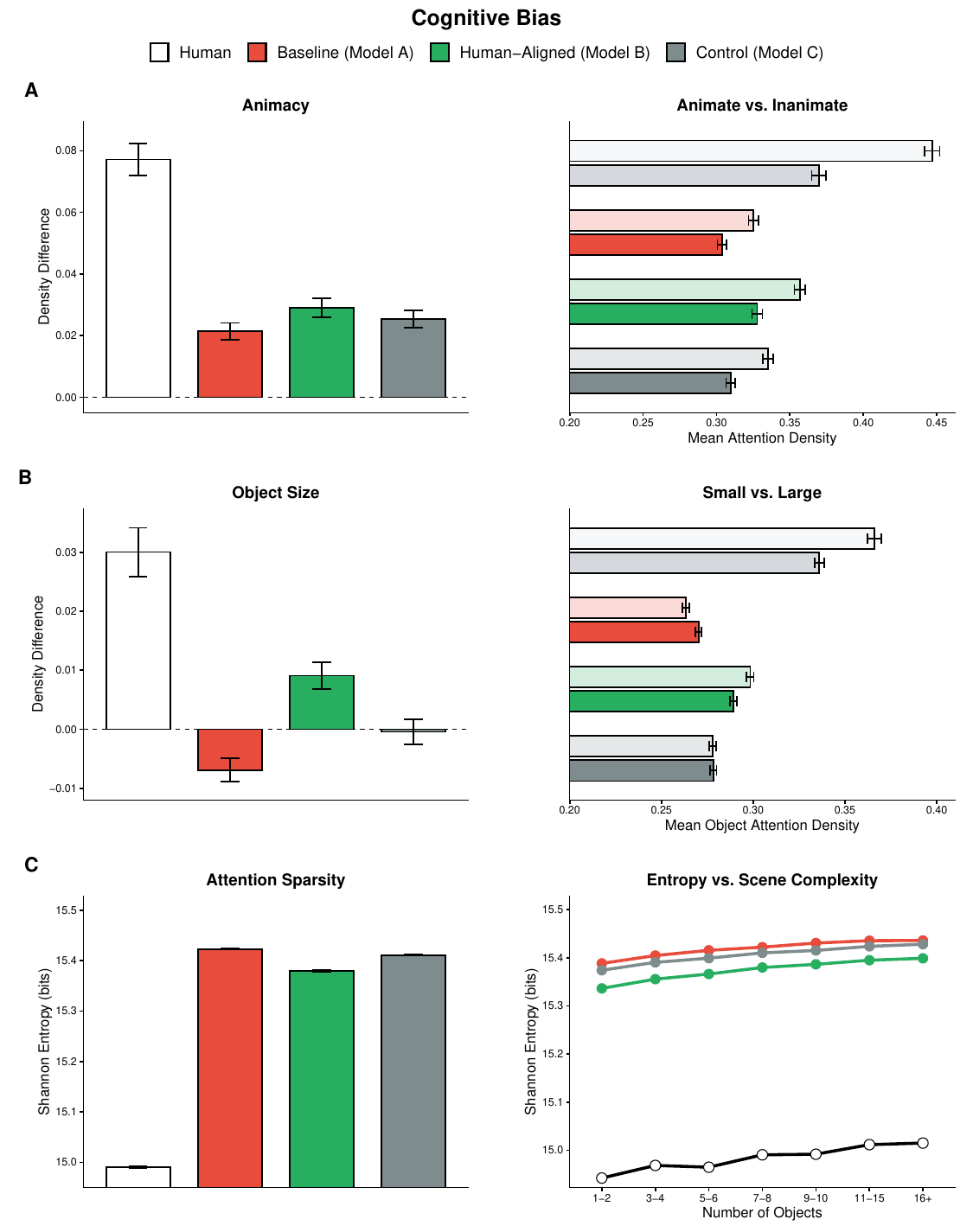}\vspace{-0.5cm}\end{figure}

\textbf{Figure 3. Acquisition of Human-Like Cognitive Biases.} Human
(white) and model (red, green, grey) biases are displayed in terms of
effect magnitude (left) and raw attention densities (right). Error bars
denote ±1 SEM. \textbf{(A) Animacy.} Fine-tuning (green) significantly
amplified the preference for animate (light) over inanimate (dark)
regions compared to the baseline (red) and shuffled control (grey).
\textbf{(B) Object Size.} The baseline model exhibited an anti-human
bias, preferentially attending to large (dark) over small (light)
objects, which was reversed by fine-tuning, inducing a reliable
human-like preference for small objects. \textbf{(C) Sparsity.}
Attention entropy (bits) served as a measure of spatial concentration.
The tuned model was significantly less entropic than the baseline and
control across all levels of scene clutter (number of annotated
objects), shifting toward the high sparsity characteristic of human
fixations.

\subsection{Benchmark Evaluation}\label{benchmark-evaluation}

Fine-tuning for human-like attention came at no cost to classification
performance across in-distribution, corrupted and out-of-distribution
benchmarks (Figure 4). Specifically, there was decisive Bayesian
evidence of parity for classification performance between the tuned and
baseline model on the ImageNet (BF\textsubscript{01} = 222) and
ImageNet-C validation sets (BF\textsubscript{01} = 242.8), showing that
the tuning procedure did not introduce vulnerability to common
distortions (e.g., JPEG compression, visual snow, fog). Likewise, very
strong evidence of Bayesian equivalence was also observed when using the
out-of-distribution ObjectNet set benchmark (BF\textsubscript{01} = 89).

\begin{figure}[H]\centering\includegraphics[width=0.9\textwidth,keepaspectratio]{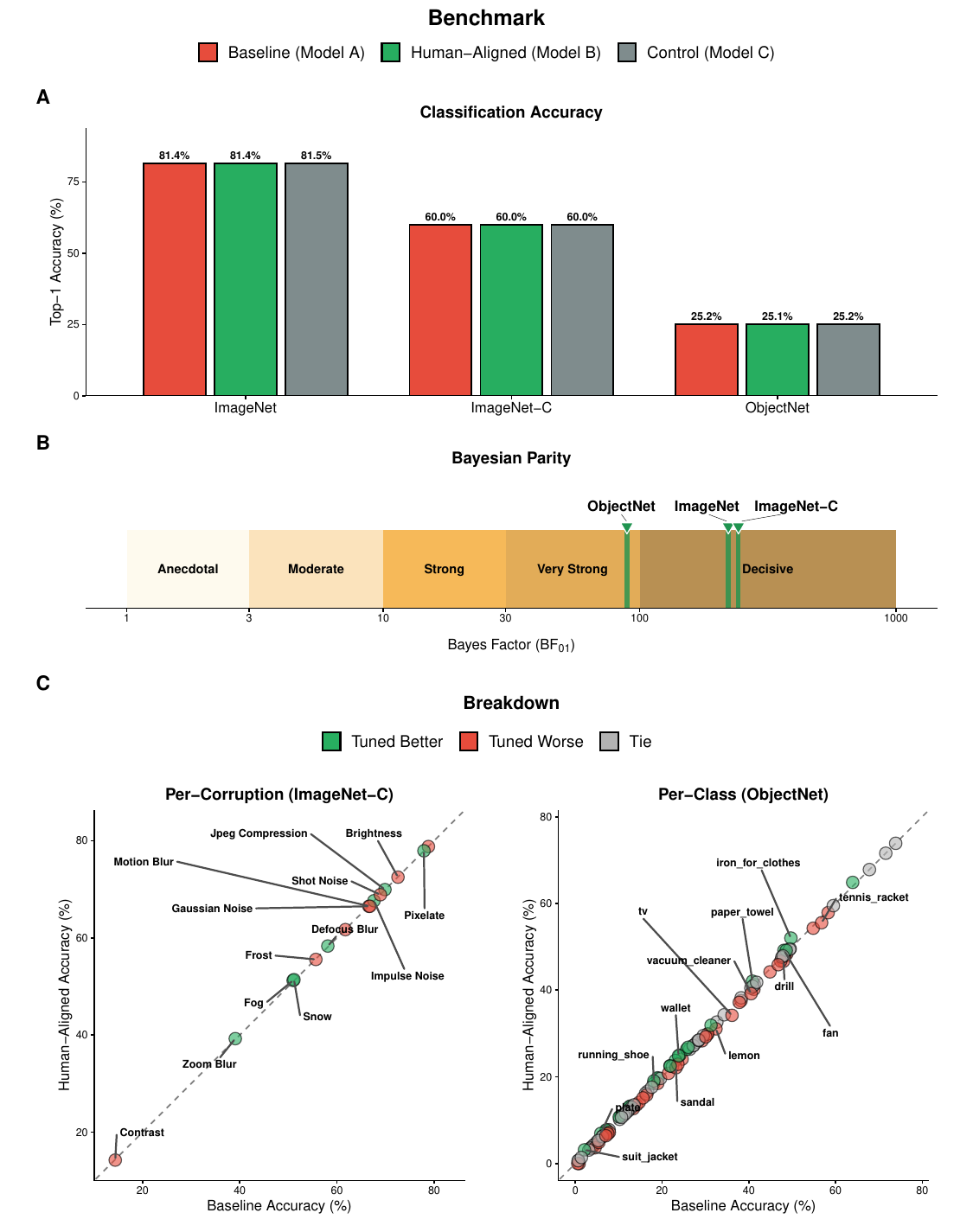}\vspace{-0.5cm}\end{figure}

\textbf{Figure 4. Benchmarks.}~\textbf{(A) Classification Accuracy.}
Top-1 accuracy on ImageNet (in-distribution), ImageNet-C (corrupted) and
ObjectNet (out-of-distribution) sets for the baseline (red), tuned
(green) and shuffled control (grey) models. \textbf{(B) Bayesian
Evidence of Equivalence.} Bayes factors for the tuned model versus
baseline on each benchmark, plotted on a log scale against
Jeffreys\textquotesingle{} evidence tiers. \textbf{(C) Directional
Accuracy.} Accuracy for the tuned versus baseline model with points
coloured by direction (green: tuned better; red: baseline better; grey:
tie) across corruption types (left) and classes (right); diagonal
represents perfect parity.

\subsection{CNN Replication}\label{cnn-replication}

Applying the equivalent fine-tuning procedure to a ResNet-50 (Figure 5A)
produced the opposite pattern. First, cognitive alignment failed, where
all saliency alignment metrics declined substantially across all five
metrics (Pearson CC: 0.39 → 0.11; NSS: 0.49 → 0.13; AUC-Judd: 0.74 →
0.58; KL-divergence: 0.90 → 1.86; SIM: 0.59 → 0.45; all p \textless{}
.001; Figure 5B), where the tuned model outperformed the baseline on
only \textasciitilde15\% of images and for a single image category
(Figure 5C). Second, classification performance degraded severely, with
Top-1 in-distribution ImageNet accuracy catastrophically falling from
80.3\% to 68.7\% (Figure 5B inset).

\begin{figure}[H]\centering\includegraphics[width=0.9\textwidth,keepaspectratio]{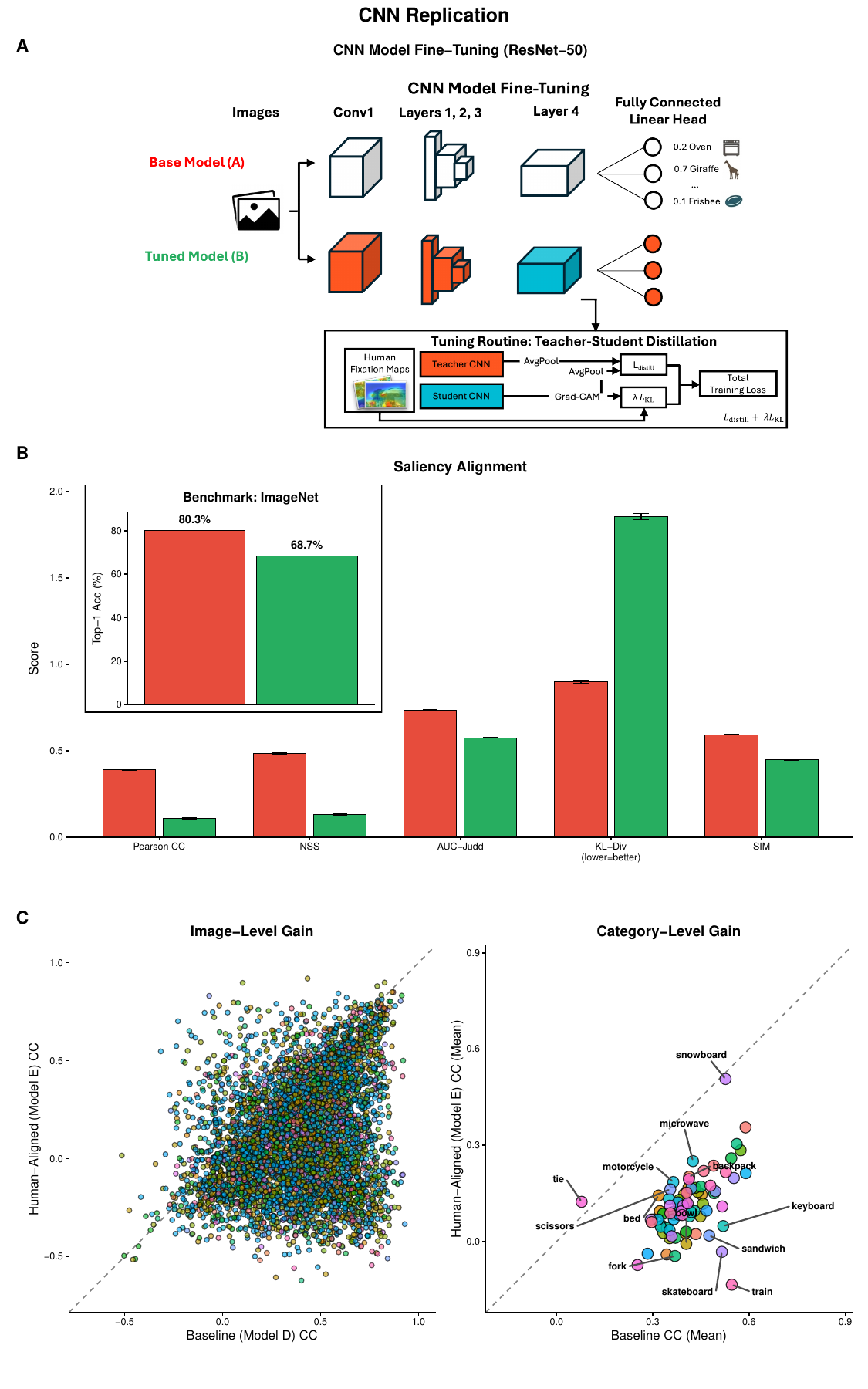}\vspace{-0.5cm}\end{figure}\textbf{Figure
5. ResNet-50 Replication.} \textbf{(A) Methods for CNN Fine-Tuning.} An
equivalent teacher-student distillation framework was used on the fourth
model layer, with Grad-CAM as the attention proxy in place of attention
rollout. \textbf{(B) Model-Human Attentional Alignment.} Alignment
metrics for the baseline (red) and tuned (green) CNN, with the benchmark
classification performance inset showing top-1 ImageNet accuracy. Unlike
the ViT, fine-tuning produced substantial declines across all alignment
(note that for KL-divergence (lower = better) and is unbounded where the
tall green bar indicates degraded performance) and benchmark metrics.
\textbf{(C) Alignment Gain.} Panels show alignment loss for the tuned
versus baseline CNN given (in contrast to Figure 2) that the points are
predominantly distributed below the diagonal.

\section{Discussion}\label{discussion}

Fine-tuning a ViT on human fixation data successfully induced three
well-established cognitive biases - preferences for animacy and
small-objects along with amplified attention sparsity - without any cost
to classification performance. Prior applications of fixation-based
fine-tuning have primarily targeted saliency prediction accuracy (e.g.,
Huang et al., 2015); the present study instead uses fine-tuning as a
tool to induce theoretically motivated cognitive biases that are
directly measurable through established psychological constructs. The
shuffled control model provides a within-experiment test of specificity:
where the control Model C also shifted marginally, the effect was
consistently weaker than the tuned Model B, confirming that it is the
image-congruent semantic structure of human fixation signals (not merely
human supervision itself) that drives the observed cognitive alignment.

These results extend a growing literature demonstrating that DNNs can be
shifted toward more human-like processing strategies via fixation-based
fine-tuning. Geirhos et al. (2018) showed that training on
Stylized-ImageNet induced shape-based representations more consistent
with human object recognition, with downstream gains in robustness.
Koorathota et al. (2023) demonstrated spatial alignment between ViT
attention maps and human fixations in a driving context using a
fixation-attention intersection loss, with accuracy gains most
pronounced under high uncertainty. The present findings are consistent
with this pattern and extend it by showing that spatial attention can be
aligned independently of feature representations, and that this
independence is quantifiable through biologically motivated metrics
rather than task accuracy alone. This is consistent with recent work
showing that aligning model representations with human perceptual
similarity does not degrade downstream performance (Muttenthaler et al.,
2025), and that distinct head clusters in self-supervised ViTs can
exhibit human-like fixation patterns without sacrificing recognition
(Yamamoto et al., 2025).

We observed decisive Bayesian evidence that the tuned model's accuracy
was preserved across the ImageNet benchmark, even when images were
visually corrupted with distortions like weather (e.g., snow) or noise
(e.g., compression) and that this extended to images outside the
training distribution (ObjectNet). Based on theory, by modifying only
self-attention weights while freezing the feed-forward sublayers, the
fine-tuning procedure dissociates attentional selection from encoded
knowledge - this mirrors the distinction in biological vision between
spatial focus and invariant object recognition (e.g., Ungerleider \&
Mishkin, 1982), effectively treating the feed-forward sublayers as an
intact fixed repository of feature representations (Geva et al., 2021)
and the attention mechanism as a tuneable filter. It is worth noting
that modern ViTs may already be less aligned with human animacy
processing than earlier CNN architectures (Diaz \& Alvarez, 2025) and
that achieving better benchmark performance does not by itself ensure
more biologically plausible strategies (Linsley et al., 2025) - both
these points motivate the need for targeted interventions of the kind
demonstrated here.

The CNN comparison sharpens this interpretation. Applying an equivalent
procedure to a ResNet-50 produced the opposite outcome: saliency
alignment declined substantially across all metrics and classification
performance degraded by 11.6 percentage points representing non-trivial
forgetting. This dissociation suggests that the success of
fixation-based fine-tuning in the ViT is not a generic property of
teacher-student distillation, but depends on the architectural
availability of a modular, separable attention mechanism. In CNNs,
attention is distributed implicitly across convolutional feature maps
rather than instantiated as explicit updatable weights, which makes it
considerably harder to apply spatial realignment without
representational damage.

Across all three cognitive bias analyses, the tuned model moved reliably
in the human-like direction but fell well short of human levels. This
modest alignment reflects a qualitative difference, rather than a mere
quantitative shortfall. Human attention is characterised by serial,
capacity-limited deployment (Duncan, 1984) that the
ViT\textquotesingle s parallel, global self-attention architecture does
not natively support and the supervision signal used here can shift
spatial biases where the architecture permits without introducing the
dynamic, recurrent prioritisation that characterises human search.
Closing this gap fully is likely to require addressing both dimensions -
Morgan et al. (2025) recently showed that a recurrent ViT trained on a
primate attention task can exhibit dynamic spatial prioritisation,
implying that recurrence is likely to be a necessary ingredient for
recovering this broader class of human-like behaviour. Both supervision-
and architecture-based approaches will be key to achieving deeper
cognitive alignment.

An important consideration is the potential circularity of the bias
findings. Since the model was trained directly on human fixation maps,
any human-like spatial pattern in the model's attention could, in
principle, reflect superficial distributional absorption rather than
genuine cognitive alignment - the model may simply have learned to mimic
where humans look without the underlying representational change that
drives those fixations in humans. Two features of the data speak against
this interpretation. First, the biases recovered here were not directly
optimised: the training loss was a pixel-level spatial divergence
applied globally, not an explicit animacy or object-size objective.
Animacy preferences and the small-object bias are emergent properties of
the fixation signal rather than explicit training targets. Second, and
more decisively, the object size result reveals a qualitative shift that
distributional absorption alone cannot explain: the baseline model did
not merely underperform humans but showed a reliably anti-human tendency
to over-attend to large objects. Fine-tuning reversed this direction
entirely, shifting the model from a significant negative bias to a
significant positive one aligned with human behaviour. This qualitative
flip - from anti-human to human-like - is difficult to attribute to
distributional absorption and suggests that the structured fixation
signal is inducing genuine representational change rather than
superficial pattern matching. A related, legitimate concern is the use
of mouse-click approximations as fixation proxies in SALICON rather than
direct eye-tracking recordings; this introduces additional noise and
potential spatial bias into the supervision signal. However, that the
fine-tuning procedure succeeded despite this imprecision arguably
strengthens the conclusion: cleaner fixation data, obtained via direct
eye-tracking at scale (e.g., from wearables) would likely yield more
pronounced alignment gains. Passive eye-tracking pipelines are making
fixation data increasingly feasible to collect at scale (Krafka et al.,
2016), and feedback alignment techniques well-established for language
modelling (Ouyang et al., 2022) offer blueprints to exploit them -
suggesting that the modest alignment demonstrated here is likely to be a
floor, not a ceiling.

The current approach to validate alignment via cognitive bias metrics
addresses common concerns with saliency map applications (Barredo
Arrieta et al., 2020; Adebayo et al., 2020) and enables a uniquely deep
sense of interpretability, where model behaviour can be understood
through the same psychological constructs that explain human gaze. Given
that these effects emerge at no cost to performance (at least for the
classification task measured here), cognitive alignment can be viewed
not as a constraint to be imposed on capable models, but as a property
that can be induced for free.

\section{Methods}\label{methods}

\subsection{Model Variants}\label{model-variants}

Three model variants were compared throughout (Figure 1A). The Base
Model (A) was Google\textquotesingle s ViT-B/16 (Dosovitskiy et al.,
2021), loaded with pretrained ImageNet weights and held entirely frozen
as the unmodified reference. The Tuned Model (B) was an identically
initialised ViT-B/16 subjected to the saliency fine-tuning procedure
described below. The Control Model (C) was produced by repeating the
identical fine-tuning procedure, but with human fixation maps
deterministically shuffled prior to training (random seed = 42), such
that each image was paired with the fixation map from a different image
in the dataset. This preserved the quantity and marginal statistical
properties of the attention supervision signal while destroying its
image-fixation correspondence with biological priors. Any divergence in
attentional behaviour between Models B and C can therefore be attributed
specifically to the structured, image-congruent content of the human
fixation signal, rather than to the generic effect of additional human
attention supervision.

All three models share the ViT-B/16 architecture: input images are
divided into non-overlapping 16$\times$16 pixel patches, linearly projected
into a 768-dimensional embedding space and processed through 12
transformer blocks where each contains a 12-head multi-head
self-attention sublayer and a feed-forward network. A prepended
{[}CLS{]} token aggregates global information across all patch tokens
and is passed to the classification head for downstream prediction.

\subsection{Image \& Human Behaviour
Dataset}\label{image-human-behaviour-dataset}

Human fixation data were drawn from the SALICON dataset (Jiang et al.,
2015), which provides fixation annotations for 10,000 training and 5,000
validation images sourced from Microsoft's COCO (Lin et al., 2014). Raw
fixation coordinates were converted to (x, y) convention and scaled
proportionally from the original image dimensions to the
model\textquotesingle s input resolution (224$\times$224 pixels). Each set of
fixation coordinates was then binned onto a spatial grid and convolved
with an isotropic Gaussian kernel ($\sigma$ = 15 pixels) to produce a
continuous density map, which was subsequently min-max normalised to the
range 0-1, yielding a probability-like distribution over image space. No
data augmentation was applied beyond the standard resizing and pixel
normalisation performed by the ViT image processor.

\subsection{Fine-Tuning Procedure}\label{fine-tuning-procedure}

Fine-tuning followed a teacher-student distillation framework (Hinton et
al., 2015; Figure 1A). The frozen Base Model (A) served as the teacher;
the student (Model B or C) was trained jointly to: (i) preserve the
teacher\textquotesingle s learned representations; and (ii) shift its
internal attention distribution toward the target fixation signal. To
minimise the risk of catastrophic forgetting, only the self-attention
layer parameters were set as trainable; patch embedding weights,
normalisation (norms) and feed-forward (multilayer perceptron; MCP)
sublayers, as well as the classification head were all frozen. This
amounts to approximately 32.7\% of total model parameters being updated
during training (the 12 multi-head self-attention modules across 12
blocks, out of the full ViT-B/16 parameter set of \textasciitilde86M).

The composite training loss at each step was:

\[L_{total} = L_{distill} + \ \lambda\  \cdot \ L_{KL}
\]

where L\textsubscript{distill}~is the mean squared error (MSE) between
the student and teacher {[}CLS{]} token hidden states at the final
transformer block, penalising representational drift from the pretrained
teacher.~\(\mathcal{L}_{\text{KL}}\)~is the Kullback-Leibler (KL)
divergence between the student\textquotesingle s attention rollout map
and the normalised human fixation density map for each image. Both terms
were weighted equally (\emph{$\lambda$}~= 1.0); no hyperparameter search was
conducted and equal weighting was adopted as a neutral default to avoid
introducing an additional tuning degree of freedom. A small epsilon (1 $\times$
10$^{-}$$^{1}$$^{0}$) was added to both distributions before computing the KL term for
numerical stability.

Training used the AdamW optimiser at a learning rate of 1 $\times$ 10$^{-}$$^{5}$ -
chosen conservatively to limit deviation from the pretrained weight
landscape, with a batch size of 8 for a maximum of 20 epochs. Evaluation
on the held-out validation subset was conducted every 200 steps. Early
stopping was applied with a patience of 3 consecutive evaluation cycles
without a minimum improvement of 0.001 in validation loss, with the best
checkpoint (lowest validation loss) restored at completion. All models
were trained on Apple Silicon (MPS backend) and their training dynamics
were monitored throughout via Weights \& Biases.

\subsection{Attention Extraction}\label{attention-extraction}

Model attention was extracted using attention rollout (Abnar \& Zuidema,
2020), selected on the basis that it accounts for the recursive
propagation of attention through residual connections across all
transformer layers, unlike last-layer or mean-layer approaches which
treat each layer independently and therefore risk misrepresenting the
cumulative flow of information to the {[}CLS{]} token. At each layer,
attention weights are averaged across all 12 heads and augmented with a
residual identity connection, then re-normalised row-wise. The resulting
per-layer matrices are multiplied sequentially from the first to the
final layer, with the {[}CLS{]} token\textquotesingle s row in the final
matrix reflecting its aggregate attention across all patch tokens. This
vector is reshaped from the 14$\times$14 patch grid to 224$\times$224 via bilinear
interpolation and min-max normalised. During fine-tuning, this procedure
was implemented in a fully differentiable manner to allow gradients to
propagate back through ~\(\mathcal{L}_{\text{KL}}\) during evaluation,
outputs were detached and converted to NumPy arrays.

\subsection{Alignment Metrics}\label{alignment-metrics}

Per-image alignment between each model\textquotesingle s rollout map and
the corresponding human fixation density map was quantified using five
established saliency metrics (Bylinskii et al., 2019): Pearson
Correlation Coefficient (CC; linear correspondence between continuous
maps), Normalised Scanpath Saliency (NSS; Peters et al., 2005; mean
normalised model attention at fixated locations), AUC-Judd (Judd et al.,
2009; area under the Receiver Operating Characteristic (ROC) curve
treating fixated pixels as positives), KL Divergence (divergence from
the model distribution to the human distribution) and Similarity (SIM;
element-wise minimum overlap between normalised distributions). These
metrics collectively capture complementary aspects of saliency
correspondence (i.e., distributional overlap, spatial precision, and
rank ordering) and are standard in the computational eye-tracking
literature. Results were aggregated across the full SALICON validation
set and summarised as mean ± standard deviation per model.

\subsubsection{Human Inter-Observer
Consistency}\label{human-inter-observer-consistency}

To contextualise alignment gains, inter-observer consistency was
computed for each image as the mean pairwise Pearson CC between
individual annotators\textquotesingle{} fixation density maps, using the
per-worker records available in the SALICON annotations. This score
quantifies the degree to which human viewers agreed on where to look,
providing an image-level estimate of target ambiguity and serving as a
proxy for the inherent difficulty of aligning model attention to human
ground truth.

First, Pearson correlations were computed between per-image consistency
and each model\textquotesingle s alignment score to assess whether model
performance simply tracks with human reliability. Second, consistency
was correlated with Model B\textquotesingle s alignment gain over the
baseline (i.e., \(\text{Pearson CC}_{B}\) - \(\text{Pearson CC}_{A}\)).
This tests whether the model\textquotesingle s improvements were robust
across all scenes (a negative or near-zero correlation), or merely
driven by "easy" images with highly unambiguous focal points (a positive
correlation).

To control for the confound that images with higher baseline alignment
have less room to improve, a headroom-normalised gain was also computed
as:

\[\text{Normalised Gain} = \frac{\text{Pearson CC}_{B} - \text{Pearson CC}_{A}}{1 - \text{Pearson CC}_{A} + \varepsilon}\]

where~\(\varepsilon = 10^{- 8}\)~prevents division by zero. Finally,
images were binned into quartiles by inter-observer consistency to
characterise how mean alignment and gain vary across the distribution.

\subsection{Cognitive Bias Analyses}\label{cognitive-bias-analyses}

Three cognitive bias analyses were conducted to assess whether
fine-tuning induced specific human-like attentional preferences: animacy
bias, object size bias and attention sparsity. Each analysis was applied
to all three models, using the same SALICON validation image set. Paired
t-tests were used to assess for differences whether within-model (e.g.,
small vs. large object sizes) or between-model (e.g., baseline vs. tuned
model) and reported with Cohen\textquotesingle s~d effect sizes.

\subsubsection{Animacy Bias}\label{animacy-bias}

To test whether fine-tuning induced preferential attention toward
animate entities, attention density was computed separately over animate
and inanimate regions of each image using COCO instance segmentation
annotations. Animate categories were defined as the 11 COCO classes
corresponding to people and animals. Five categories were excluded
entirely from analysis to avoid classification ambiguity: televisions,
laptops and mobile phones (which may depict animate content on-screen),
teddy bears (anthropomorphic form likely to attract animate-like
fixation despite being inanimate) and potted plants (biologically living
but not animate in the cognitive or attentional sense). Images
containing any excluded category were dropped in full to prevent
contextual contamination. Only images containing at least one animate
and at least one inanimate object were retained, to enable a
within-image preference comparison.

Pixel-level segmentation masks were decoded from COCO polygon and
Run-Length-Encoding (RLE) annotations using pycocotools. Where multiple
objects overlapped spatially, a painter\textquotesingle s algorithm was
applied: annotations were sorted by area in descending order and painted
sequentially onto a master canvas at the original image resolution, such
that smaller foreground objects overwrote larger background objects. The
resulting canvas was down-sampled to 224$\times$224 using nearest-neighbour
interpolation to match the attention map resolution. Each pixel was
assigned exclusively to animate, inanimate or background. Attention
density for each region was computed as the sum of normalised attention
within that region divided by its pixel area, scaled by 10,000 for
readability.

\subsubsection{Object Size Bias}\label{object-size-bias}

To test whether fine-tuning shifted attention toward smaller,
informationally dense objects, attention density was computed per
annotated object using COCO instance segmentation masks decoded
identically to the animacy analysis. Each object was classified into one
of three size bins using the COCO standard area thresholds applied to
the object\textquotesingle s pixel area in the original image: small
(\textless{} 1,024 px$^{2}$), medium (1,024--9,216 px$^{2}$) and large ($\ge$ 9,216
px$^{2}$). Attention density per object was computed as normalised attention
summed over the object\textquotesingle s resized mask, divided by mask
area and scaled by 10,000. Per-image metrics were obtained by averaging
object-level densities within each size bin. Images with fewer than two
annotated objects were excluded.

\subsubsection{Attention Sparsity}\label{attention-sparsity}

To test whether fine-tuning produced more concentrated, human-like
attention distributions, Shannon entropy was computed for each
model\textquotesingle s attention rollout map per image:

\[H = - \sum_{i}^{}{p(i)}{\log}_{2}p(i)
\]

where~\(p(i)\)~is the normalised attention value at pixel~\(i\),
obtained by flattening the 224$\times$224 map to a 1D vector, clipping negative
values and normalising to sum to one. Entropy is expressed in bits;
lower values indicate more spatially concentrated attention.

\subsection{Benchmark: Classification Performance
Evaluation}\label{benchmark-classification-performance-evaluation}

To assess whether fine-tuning came at any cost to the
models\textquotesingle{} original classification ability, downstream
performance was evaluated across three benchmarks. For standard
preprocessing across all datasets, images were resized to 256$\times$256,
centre-cropped to 224$\times$224 and normalised to a mean and standard
deviation of 0.5 per channel, consistent with the ViT-B/16
specification.

The first benchmark evaluated in-distribution performance using the
standard ImageNet-1k validation set (50,000 images across 1,000
classes). Second, robustness against common visual distortions was
assessed using the ImageNet-C benchmark (Hendrycks \& Dietterich, 2019),
incorporating all 15 official corruption types (spanning noise, blur,
weather, and digital perturbations) at severity level 5. To ensure
methodological correctness, corruptions were applied dynamically
in-memory to a 5,000-image subset of the standard validation dataset
using the imagecorruptions library; crucially, corruptions were injected
into unnormalised RGB images prior to the standard ViT preprocessing
steps. Finally, generalisation to out-of-distribution inputs was tested
using ObjectNet (Barbu et al., 2019), which controls for viewpoint,
rotation and background. Following the standard ObjectNet evaluation
protocol, only images mapping to valid ImageNet-1k indices were included
(\textasciitilde18,500 images across 113 classes). For ImageNet-1k and
ImageNet-C, Top-1 accuracy was recorded per image and aggregated. Due to
the one-to-many label mapping inherent to ObjectNet, a prediction on
this dataset was scored as correct if the Top-1 predicted class matched
any valid ImageNet label index associated with that specific image.

\subsubsection{Bayesian Parity Analysis}\label{bayesian-parity-analysis}

Given the large sample sizes involved, even negligible differences in
accuracy would yield statistically significant frequentist tests. To
instead quantify evidence in favour of performance equivalence between
Model A and Model B, a Bayes factor BF$_{0}$$_{1}$ was computed for all benchmarks
using the Jeffreys-Zellner-Siow approximation applied to the
paired~t-statistic on per-image top-1 accuracy. A BF$_{0}$$_{1}$ \textgreater{} 1
indicates evidence favouring the null hypothesis of no difference
between models; values are interpreted on Jeffreys\textquotesingle{}
scale (anecdotal: 1-3; moderate: 3-10; strong: 10-30; very strong:
30-100; decisive \textgreater100; Rouder et al., 2009) to directly tests
the "at no cost" claim central to this study.

\subsection{Generalisability to CNN
Architectures}\label{generalisability-to-cnn-architectures}

To determine whether the effects of attention fine-tuning generalise
beyond a ViT, the experiment was replicated using a ResNet-50 backbone
(He et al., 2016). We employed the same teacher-student distillation
framework with two primary architectural adaptations. First, as CNNs
lack the global, recursive self-attention mechanism, we extracted
spatial attention via Grad-CAM (Selvaraju et al., 2017). Grad-CAM
generates spatial activation maps by weighting the feature maps of a
target convolutional layer by the global average-pooled gradients of the
target class.

Second, to mitigate representational drift while facilitating meaningful
updates, we unfroze only the final convolutional block (layer4); the
preceding stages (conv1 through layer3) and the classification head
remained frozen. The distillation loss was computed as the MSE between
the student and teacher outputs at the global average pooling layer
(represented as a 2048-dimensional vector) substituting for the ViT
{[}CLS{]} token. To account for the magnitude differences between
Grad-CAM outputs and ViT attention rollout, the attention loss scalar
was tuned to ($\lambda$ = 0.01). The learning rate was halved to 5 $\times$
10\^{}\textsuperscript{-6} to ensure numerical stability.

\newpage
\section{References}\label{references}

Adebayo, J., Gilmer, J., Muelly, M., Goodfellow, I., Hardt, M., \& Kim,
B. (2020). Sanity checks for saliency maps.
https://doi.org/10.48550/arXiv.1810.03292

Barredo Arrieta, A., Díaz-Rodríguez, N., Del Ser, J., Bennetot, A.,
Tabik, S., Barbado, A., Garcia, S., Gil-Lopez, S., Molina, D.,
Benjamins, R., Chatila, R., \& Herrera, F. (2020). Explainable
Artificial Intelligence (XAI): Concepts, taxonomies, opportunities and
challenges toward responsible AI. Information Fusion, 58, 82--115.

Bi, J., Guo, J., Tang, Y., Wen, L. B., Liu, Z., Wang, B., \& Xu, C.
(2025). Unveiling visual perception in language models: An attention
head analysis approach. In~\emph{Proceedings of the Computer Vision and
Pattern Recognition Conference}~(pp. 4135-4144).

Bruce, N. D., \& Tsotsos, J. K. (2009). Saliency, attention, and visual
search: An information theoretic approach. Journal of vision, 9(3), 5-5.

Bylinskii, Z., Judd, T., Oliva, A., Torralba, A., \& Durand, F. (2019).
What do different evaluation metrics tell us about saliency models?
\emph{IEEE Transactions on Pattern Analysis and Machine Intelligence,
41}(3), 740-757.

Carrasco, M., González-Martín, C., Aranda, J., \& Oliveros, L. (2025).
Vision Transformer attention alignment with human visual perception in
aesthetic object evaluation. arXiv preprint arXiv:2507.17616.

Choi, M., Zhang, Y., Han, K., Wang, X., \& Liu, Z. (2024). Human
Eyes--Inspired Recurrent Neural Networks Are More Robust Against
Adversarial Noises. Neural Computation, 36(9), 1713-1743.

Deng, J.; Dong, W.; Socher, R.; Li, L.-J.; Li, K.; and Fei-Fei, L. 2009.
Imagenet: A large-scale hierarchical image database. In Proceedings of
the IEEE Conference on Computer Vision and Pattern Recognition,
248--255.

Diaz, L. T., \& Alvarez, G. A. Ventral Stream Responses to Inanimate
Objects are Equally Aligned with AlexNet (2012) and Modern Deep Neural
Networks.
\url{https://2025.ccneuro.org/abstract_pdf/Diaz_2025_Ventral_Stream_Responses_Inanimate_Objects_Equally.pdf}

Dosovitskiy, A., Beyer, L., Kolesnikov, A., Weissenborn, D., Zhai, X.,
Unterthiner, T., Dehghani, M., Minderer, M., Heigold, G., Gelly, S.,
Uszkoreit, J., \& Houlsby, N. (2021). An image is worth 16x16 words:
Transformers for image recognition at scale.
https://doi.org/10.48550/arXiv.2010.11929

Duncan, J. (1984). Selective attention and the organization of visual
information.~\emph{Journal of experimental psychology:
General},~\emph{113}(4), 501.

Frischen, A., Bayliss, A. P., \& Tipper, S. P. (2007). Gaze cueing of
attention: visual attention, social cognition, and individual
differences.~\emph{Psychological bulletin},~\emph{133}(4), 694.

Geirhos, R., Rubisch, P., Michaelis, C., Bethge, M., Wichmann, F. A., \&
Brendel, W. (2018, November). ImageNet-trained CNNs are biased towards
texture; increasing shape bias improves accuracy and robustness. In
International conference on learning representations.

Geva, M., Schuster, R., Berant, J., \& Levy, O. (2021, November).
Transformer feed-forward layers are key-value memories. In Proceedings
of the 2021 Conference on Empirical Methods in Natural Language
Processing (pp. 5484-5495).

Hayhoe M. M., Rothkopf C. A. (2011). Vision in the natural world. Wiley
Interdiscip. Rev. Cogn. Sci. 2 158--166 10.1002/wcs.113

He, K., Zhang, X., Ren, S., \& Sun, J. (2016). Deep residual learning
for image recognition. In Proceedings of the IEEE conference on computer
vision and pattern recognition (pp. 770-778).

Hermann, K., Chen, T., \& Kornblith, S. (2020). The origins and
prevalence of texture bias in convolutional neural networks. Advances in
neural information processing systems, 33, 19000-19015.

Henderson, J. M., \& Hayes, T. R. (2017). Meaning-based guidance of
attention in scenes as revealed by meaning maps.~\emph{Nature human
behaviour},~\emph{1}(10), 743-747.

Hendrycks, D., \& Dietterich, T. (2019). Benchmarking neural network
robustness to common corruptions and perturbations. arXiv preprint
arXiv:1903.12261.

Hinton, G., Vinyals, O., \& Dean, J. (2015). Distilling the knowledge in
a neural network. arXiv preprint arXiv:1503.02531.

Huang, X., Shen, C., Boix, X., \& Zhao, Q. (2015). Salicon: Reducing the
semantic gap in saliency prediction by adapting deep neural networks. In
Proceedings of the IEEE international conference on computer vision (pp.
262-270).

Itti, L., \& Baldi, P. (2009). Bayesian surprise attracts human
attention.~\emph{Vision research},~\emph{49}(10), 1295-1306.

Itti, L., \& Koch, C. (2001). Computational modelling of visual
attention. Nature reviews neuroscience, 2(3), 194-203.

Janiszewski, C. (1998). The Influence of display characteristics on
visual exploratory search behavior. J. Consum. Res. 25, 290--301. doi:
10.1086/209540

Jiang, M., Huang, S., Duan, J., \& Zhao, Q. (2015). Salicon: Saliency in
context. In Proceedings of the IEEE conference on computer vision and
pattern recognition (pp. 1072-1080).

Judd, T., Ehinger, K., Durand, F., \& Torralba, A. (2009, September).
Learning to predict where humans look. In 2009 IEEE 12th international
conference on computer vision (pp. 2106-2113). IEEE.

Kanwisher, N., \& Yovel, G. (2006). The fusiform face area: a cortical
region specialized for the perception of faces. Philosophical
Transactions of the Royal Society B: Biological Sciences, 361(1476),
2109-2128.

Knights, E., Mansfield, C., Tonin, D., Saada, J., Smith, F. W., \&
Rossit, S. (2021). Hand-selective visual regions represent how to grasp
3D tools: Brain decoding during real actions. Journal of Neuroscience,
41(24), 5263-5273.

Knights, E., Smith, F. W., \& Rossit, S. (2022). The role of the
anterior temporal cortex in action: evidence from fMRI multivariate
searchlight analysis during real object grasping. Scientific Reports,
12(1), 9042.

Koorathota, S., Papadopoulos, N., Ma, J. L., Kumar, S., Sun, X., Mittal,
A., ... \& Sajda, P. (2023). Fixating on attention: Integrating human
eye tracking into vision transformers. arXiv preprint arXiv:2308.13969.

Krajbich I., Armel C., Rangel A. (2010). Visual fixations and the
computation and comparison of value in simple choice. Nat. Neurosci. 13
1292--1298 10.1038/nn.263

Kirillov, A., Mintun, E., Ravi, N., Mao, H., Rolland, C., Gustafson, L.,
... \& Girshick, R. (2023). Segment anything. In~\emph{Proceedings of
the IEEE/CVF international conference on computer vision}~(pp.
4015-4026).

Krizhevsky, A., Sutskever, I., \& Hinton, G. E. (2012). Imagenet
classification with deep convolutional neural networks. Advances in
neural information processing systems, 25.

LeCun, Y., \& Bengio, Y. (1998). Convolutional networks for images,
speech, and time series. The handbook of brain theory and neural
networks.

Li, X., Yin, X., Li, C., Zhang, P., Hu, X., Zhang, L., ... \& Gao, J.
(2020, August). Oscar: Object-semantics aligned pre-training for
vision-language tasks. In~\emph{European conference on computer
vision}~(pp. 121-137). Cham: Springer International Publishing.

Lin, T. Y., Maire, M., Belongie, S., Hays, J., Perona, P., Ramanan, D.,
... \& Zitnick, C. L. (2014, September). Microsoft coco: Common objects
in context. In~\emph{European conference on computer vision}~(pp.
740-755). Cham: Springer International Publishing.

Linsley, D., Feng, P., \& Serre, T. (2025). Better artificial
intelligence does not mean better models of biology. Trends in Cognitive
Sciences. https://doi.org/10.1016/j.tics.2025.11.016

Lu, X., \& Jiang, Y. (2026). How Visual Experience Shapes Face
Processing: Divergent Representational Strategies Emerge from
Specialized and Diverse Visual Diets in Artificial Neural Networks.
bioRxiv, 2026-02.

Ungerleider, L. \& Mishkin, M. (1982). Two cortical visual systems. In

Analysis of Visual Behavior (Ingle, D. et al. Eds.), pp. 549--586, MIT
Press

Morgan, J., Albanna, B. F., \& Herman, J. P. (2025). A recurrent vision
transformer shows signatures of primate visual attention.
arXiv:2502.10955.

New, J., Cosmides, L., \& Tooby, J. (2007). Category-specific attention
for animals reflects ancestral priorities, not expertise. Proceedings of
the National Academy of Sciences, 104(42), 16598-16603.

Nuthmann, A., \& Henderson, J. M. (2010). Object-based attentional
selection in scene viewing.~\emph{Journal of vision},~\emph{10}(8),
20-20.

Ouyang, L., Wu, J., Jiang, X., Almeida, D., Wainwright, C., Mishkin, P.,
... \& Lowe, R. (2022). Training language models to follow instructions
with human feedback. Advances in neural information processing systems,
35, 27730-27744.

Parkhurst D., Law K., Niebur E. (2002). Modeling the role of salience in
the allocation of overt visual attention. Vision Res. 42 107--123
10.1016/S0042-6989(01)00250-4

Pegler, D., Steyrl, D., Zhang, M., Karner, A., Arato, J., Scharnowski,
F., \& Melinscak, F. (2025). SpiderNets: Vision Models Predict Human
Fear From Aversive Images.~\emph{arXiv preprint arXiv:2509.04889}.

Peschel, A. O., \& Orquin, J. L. (2013). A review of the findings and
theories on surface size effects on visual attention. Frontiers in
psychology, 4, 902.

Peters, R. J., Iyer, A., Itti, L., \& Koch, C. (2005). Components of
bottom-up gaze allocation in natural images. Vision research, 45(18),
2397-2416.

Petersen, S. E., \& Posner, M. I. (2012). The attention system of the
human brain: 20 years after.~\emph{Annual review of
neuroscience},~\emph{35}, 73-89.

Piñero, L. G. O., Carrasco, M., Aranda, J., \& González, C. (2025).
Comparative perspective of visual attention: From human focus to visual
transformers --- an in-depth review. \emph{Cognition}, \emph{20},
172230.

Radford, A., Kim, J. W., Hallacy, C., Ramesh, A., Goh, G., Agarwal, S.,
... \& Sutskever, I. (2021, July). Learning transferable visual models
from natural language supervision. In~\emph{International conference on
machine learning}~(pp. 8748-8763). PmLR.

Selvaraju, R. R., Cogswell, M., Das, A., Vedantam, R., Parikh, D., \&
Batra, D. (2017). Grad-CAM: Visual explanations from deep networks via
gradient-based localization. Proceedings of the IEEE International
Conference on Computer Vision (ICCV), 618--626.

Tatler, B. W. (2007). The central fixation bias in scene viewing:
Selecting an optimal viewing position independently of motor biases and
image feature distributions.~\emph{Journal of vision},~\emph{7}(14),
4-4.

Treisman, A. M., \& Gelade, G. (1980). A feature-integration theory of
attention. Cognitive psychology, 12(1), 97-136.

Valenza, E., Simion, F., Cassia, V. M., \& Umiltà, C. (1996). Face
preference at birth.~\emph{Journal of experimental psychology: Human
Perception and Performance},~\emph{22}(4), 892.

Vaswani A., Shazeer N., Parmar N., Uszkoreit J., Jones L., Gomez A.N.,
Kaiser L.u., Polosukhin I. Attention is All you Need. In: Guyon I.,
Luxburg U.V., Bengio S., Wallach H., Fergus R., Vishwanathan S., Garnett
R., editors. Advances in Neural Information Processing Systems. Vol. 30
Curran Associates, Inc.; Red Hook, NY, USA: 2017

Walther, D., \& Koch, C. (2006). Modeling attention to salient
proto-objects.~\emph{Neural networks},~\emph{19}(9), 1395-1407.

Yamamoto, T., Akahoshi, H., \& Kitazawa, S. (2025). Emergence of
human-like attention and distinct head clusters in self-supervised
vision transformers: A comparative eye-tracking study. Neural Networks,
189, 107595.

Yamins, D. L., \& DiCarlo, J. J. (2016). Using goal-driven deep learning
models to understand sensory cortex. Nature neuroscience, 19(3),
356-365.

\end{document}